\documentclass[letterpaper]{article} 
\usepackage[preprint]{aaai2027}  
\usepackage[hyphens]{url}  
\usepackage{graphicx} 
\usepackage{natbib}  
\usepackage{caption} 
\usepackage{algorithm}
\usepackage{algorithmic}

\usepackage{newfloat}
\usepackage{listings}
\DeclareCaptionStyle{ruled}{labelfont=normalfont,labelsep=colon,strut=off} 
\floatstyle{ruled}
\newfloat{listing}{tb}{lst}{}
\floatname{listing}{Listing}

\usepackage{booktabs}

\usepackage{multirow}
\usepackage{makecell}
\usepackage{amsmath}
\usepackage{amssymb}

\title{ReXMo: End-to-End Video Character Replacement with Expression and Motion Preservation}

\author{
    Hualiang Wei\textsuperscript{\rm 1},
    Wenhui Li\textsuperscript{\rm 1}\corresponding
}

\affiliations{
    \textsuperscript{\rm 1}College of Computer Science and Technology, Jilin University\\
    Changchun, Jilin, China
}

\begin{document}

\makeatletter
\let\originalmaketitle\@maketitle
\renewcommand{\@maketitle}{%
  \originalmaketitle

  \centering
  \includegraphics[width=0.95\linewidth]{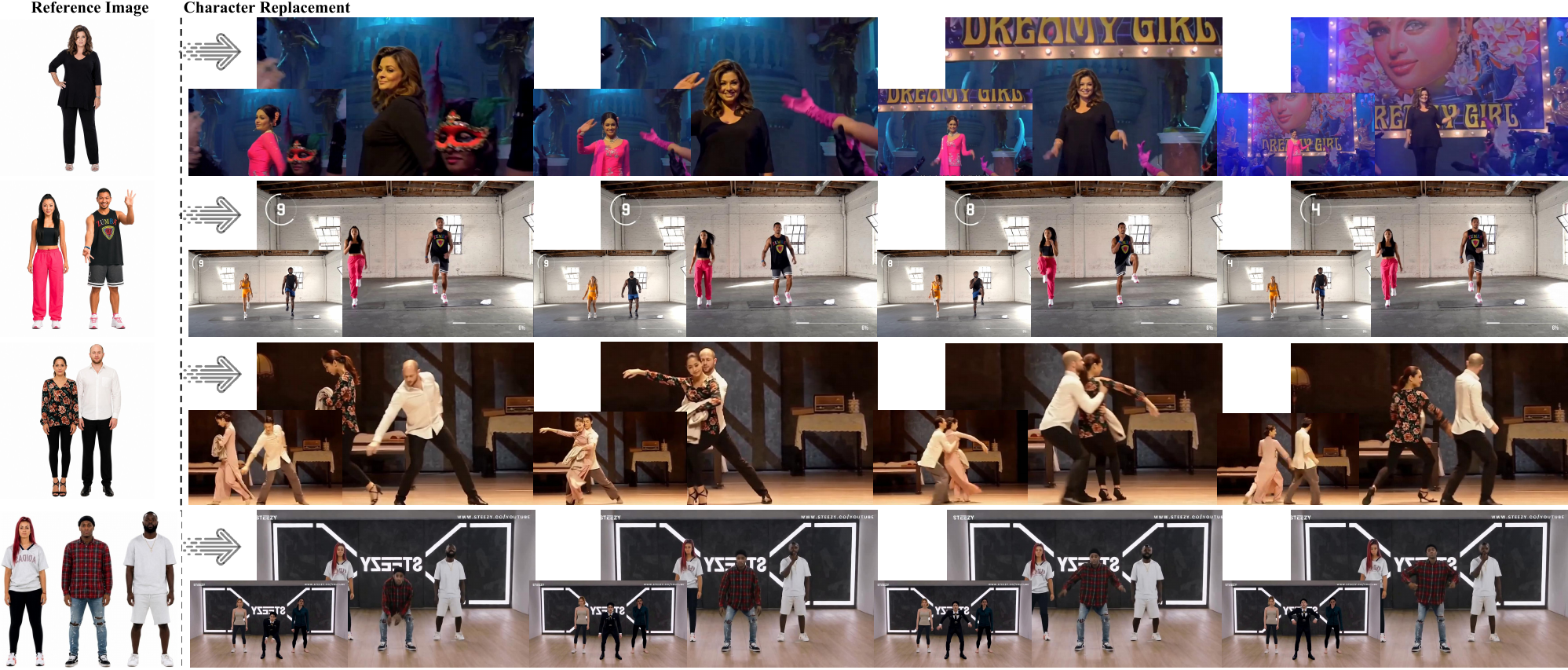}

  \captionof{figure}{
   We propose ReXMo, a video character replacement framework that enables high-fidelity identity transformation while preserving facial expressions, motions, and scene dynamics under diverse and challenging scenarios, including large motion variations, stylized identities, and multi-character interactions.
  }
  \label{fig:teaser}

  \bigskip
}
\makeatother

\maketitle

\begin{abstract}
Video character replacement aims to replace actors while preserving facial expressions, motions, and scene dynamics. Existing methods typically rely on masked generation with pose guidance, limiting fine-grained expression and motion preservation in complex scenes. We propose \textbf{ReXMo}, an end-to-end framework that supports single-, multi-, and selective character replacement scenarios. ReXMo introduces paired triplet supervision, where synthesized driving videos and original videos share identical performances but different identities, enabling scalable training through synthetic data generation. An In-Context Diffusion Transformer learns identity transformation by jointly conditioning on driving videos, reference identities, and text prompts, while an Expression Enhancement Adapter improves facial expression preservation in multi-person scenes. Extensive experiments demonstrate that ReXMo outperforms existing methods in identity fidelity, motion consistency, expression preservation, and visual quality. Our dataset and code will be publicly released.
\end{abstract}


\section{1. Introduction}

Character replacement in videos is a fundamental capability for film production, visual effects, virtual actors, and personalized content creation~\cite{liu2019neural,wang2024replace,zhang2025generative,xu2026mocha}. Given an input video and one or more reference identities, the goal is to replace the target actors while faithfully preserving the original performance, camera motion, scene layout, actor-object interactions, and temporal dynamics. To be practically useful, a practical character replacement framework should be capable of handling single-character, multi-character, and selective replacement scenarios.

Despite the progress of diffusion-based image and video generation~\cite{rombach2022high,peebles2023scalable,ho2022video,wan2025wan,brooks2024video}, high-fidelity video character replacement remains challenging. Existing methods~\cite{wang2024replace,wei2025mocha,yan2026scail,yan2026scail2,xu2024magicanimate,zhang2024mimicmotion,hu2026multianimate} usually formulate the task as masked generation or pose-guided video generation, where explicit masks and motion representations, such as skeletons or dense poses, are used to guide character regeneration from reference identities. This design is largely motivated by the scarcity of paired replacement videos with identical performances but different identities, forcing prior methods to rely on indirect supervision.

However, such a formulation has several inherent limitations. First, pose or skeleton guidance~\cite{guler2018densepose,yang2023effective,cao2017realtime,hu2024animate} provides only a sparse and abstract description of human motion. While they~\cite{wang2024replace,xu2024magicanimate,zhang2024mimicmotion} can capture coarse body structure, it often fails to preserve fine-grained performance details, including facial expressions, hand motions, and subtle body dynamics. These details are crucial for realistic character replacement, especially in cinematic videos where emotional expression and acting style are as important as global motion. Second, mask-based replacement heavily depends on accurate human segmentation. Even advanced segmentation models~\cite{cheng2022masked,ravi2025sam,jain2023oneformer,zou2023segment,ke2023segment} can produce unstable or inaccurate masks under occlusions, crowded scenes, and complex backgrounds. These errors may cause artifacts, temporal flickering, and inconsistent actor-scene interactions, especially in multi-person scenarios.


To overcome these limitations, we revisit the data formulation for video character replacement. Instead of learning from masked reconstruction or pose-only guidance, we propose to learn from paired triplet supervision. Each triplet consists of a synthesized driving video, an original video, and a reference identity condition, where the two videos share identical motions, facial performances, and scene contexts but differ in character identities. This formulation directly supervises identity transformation while preserving the original performance and video dynamics. Since such paired data is rarely available in real-world datasets, we construct it synthetically. Starting from existing human videos, we progressively replace individual characters using a strong character animation and replacement pipeline, producing aligned video pairs with controlled identity changes. By applying this process to single actors, multiple actors, and selected actors in multi-person scenes, we obtain triplet data for single-character, multi-character, and selective character replacement.

Based on this paired triplet supervision, we introduce an in-context Diffusion Transformer framework for video character replacement. We formulate character replacement as an in-context video-to-video transformation, where the model learns identity transformation while attending to the original performance. Specifically, the synthesized driving video latent is concatenated with the noisy original latent through channel-wise concatenation, while the reference identity condition is introduced through in-context token concatenation. This unified conditioning enables the diffusion transformer to establish rich spatial-temporal correspondences between the driving performance and the reference identity, allowing the model to preserve subtle motion patterns, actor-object interactions, background consistency, and temporal coherence while transforming character identity.


In addition, we introduce an Expression Enhancement Adapter to improve facial expression preservation in multi-person scenarios. Facial details are particularly challenging to maintain when actors appear at small scales, since latent video representations may weaken subtle expression variations in compressed facial regions. To address this issue, the proposed adapter extracts expression representations from temporally aligned facial sequences and injects them into the diffusion transformer through Face Adapter modules. By enhancing facial features independently for different actors, the adapter prevents expression interference and improves expression preservation for single-, multi-, and selective character replacement.

As shown in Fig.~\ref{fig:teaser}, our framework supports a unified set of character replacement modes, including single-, multi-, and selective character replacement. These capabilities enable flexible identity transformation in complex multi-person scenes while preserving the original performance and scene dynamics.

Our contributions are summarized as follows:

\begin{itemize}
    \item We introduce paired triplet supervision for video character replacement. Instead of relying on masked reconstruction or pose-only guidance, our method learns from aligned synthesized driving videos and original videos that share identical performances but different identities, together with reference identity and text conditions.

    \item We propose an in-context Diffusion Transformer that supports diverse character replacement scenarios, including single-, multi-, and selective replacement. By jointly conditioning on synthesized driving videos, reference identities, and text prompts, our model learns identity transformation while preserving motion, expressions, actor interactions, and scene consistency.

    \item We design an Expression Enhancement Adapter for fine-grained facial expression preservation in multi-person scenes. We further construct a scalable synthetic paired-data generation pipeline and will release the resulting dataset and code to facilitate future research.
\end{itemize}

\begin{figure*}[t]
\centering
\includegraphics[width=0.95\textwidth]{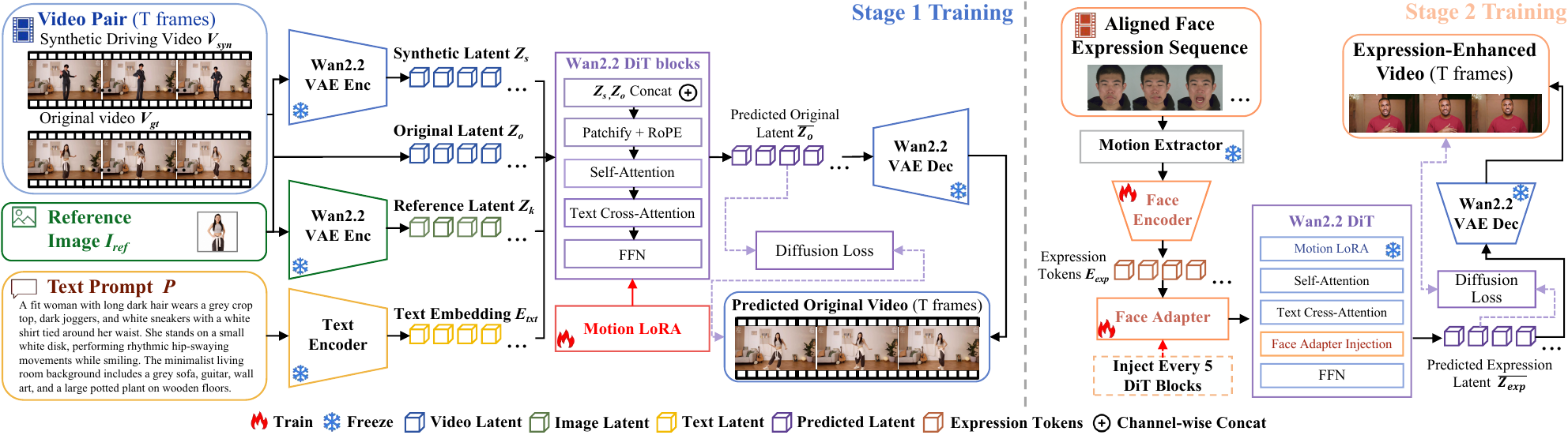} 
\caption{Overview of the ReXMo framework. Stage I learns reference-guided character replacement with synthetic cross-identity paired triplet supervision. Stage II freezes the replacement model and optimizes the Expression Enhancement Adapter with aligned facial sequences for fine-grained expression preservation.}
\label{pipleline}
\end{figure*}

\section{2. Related Work}

\subsection{2.1 Multi-character Animation}

Recent studies~\cite{zhu2024champ,tan2024animate,wang2024vividpose,tu2025stableanimator,cheng2025wan,jiang2025vace} have extended controllable animation from single-person synthesis to multi-character scenarios. DanceTogether~\cite{chen2025dancetogether} and Follow-Your-Pose V2~\cite{xue2024follow} exploit masks, depth, and optical flow to model spatial interactions among multiple characters. MultiAnimate~\cite{hu2026multianimate} introduces Identifier Assigner and Identifier Adapter to maintain identity consistency among multiple interacting subjects, while SCAIL~\cite{yan2026scail} proposes a 3D-consistent pose representation with an in-context pose injection mechanism to improve complex motion understanding. More recently, SCAIL-2~\cite{yan2026scail2} eliminates explicit pose representations and formulates character animation as an end-to-end in-context generation problem, achieving strong generalization across diverse animation scenarios.


Despite these advances, existing approaches~\cite{xu2026mocha,wang2024replace,yan2026scail,hu2026multianimate} mainly focus on controllable character animation and often rely on intermediate motion representations, such as poses, masks, or other structural conditions. Such representations provide effective motion guidance but may lose fine-grained performance details, including facial expressions, subtle body movements, actor-object interactions, and complex occlusion patterns. These limitations become more prominent when extending character animation techniques to practical video character replacement scenarios involving multiple identities and complex interactions.


\subsection{2.2 End-to-end Character Replacement}

To reduce the information loss caused by explicit intermediate representations, recent works~\cite{luo2025dreamactor,luo2026dreamactor,cheng2025wan,jiang2025vace,xu2026mocha} explore end-to-end character replacement. DreamActor-M2~\cite{luo2026dreamactor} improves motion transfer through end-to-end alignment, MoCha~\cite{xu2026mocha} leverages synthetic data for replacement training, and SCAIL-2~\cite{yan2026scail2} adopts in-context conditioning with driving videos.

However, these methods mainly target general animation or single-character replacement. Identity conditioning and facial enhancement methods such as DirectSwap~\cite{wang2025directswap} and Slot-ID~\cite{lai2026slot} improve complementary aspects but are not designed for full-body replacement. ReXMo addresses this gap through paired triplet supervision and an Expression Enhancement Adapter, enabling high-fidelity character replacement while preserving motion, identity, and expression consistency.



\section{3. Methods}

\begin{figure*}[t]
\centering
\includegraphics[width=0.95\textwidth]{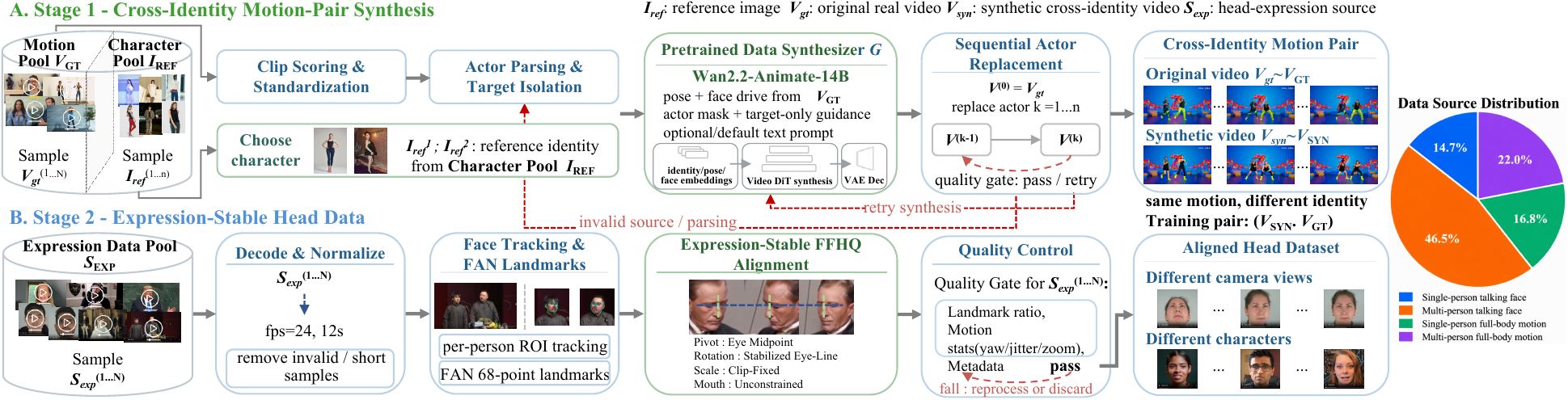} 
\caption{Overview of the proposed synthetic data generation pipeline. Stage I constructs cross-identity motion pairs for base-model training by progressively replacing target identities while preserving motions and scene dynamics. Stage II builds an expression-stable head dataset for training the Expression Enhancement Adapter through face alignment and quality filtering.}
\label{data}
\end{figure*}

\subsection{3.1 Overall Architecture}

Given an input video, one or multiple reference identity images, and a text prompt, ReXMo performs video character replacement through a decoupled two-stage training framework, as shown in Fig.~\ref{pipleline}. Stage I learns reference-guided character replacement from synthetic paired triplet supervision, while Stage II freezes the Stage-I replacement model and optimizes an Expression Enhancement Adapter for fine-grained facial expression preservation.

In Stage I, the synthesized driving video, original video, the reference identity image, and the text prompt are encoded into latent representations by the frozen Wan2.2 encoders. The diffusion backbone is then parameter-efficiently adapted using Motion LoRA to learn cross-identity character replacement while preserving motion, scene dynamics, and actor-object interactions through paired supervision.

In Stage II, the replacement model learned in Stage I remains frozen. Aligned facial expression sequences are first processed by a frozen Motion Extractor to extract temporally consistent facial motion features, which are further encoded by a lightweight Face Encoder into compact expression tokens. These tokens are injected into the frozen diffusion backbone through Face Adapter modules, enabling accurate facial expression transfer while preserving the identity replacement capability learned in Stage I.

\subsection{3.2 Controllable Video Editing Framework}

We formulate video character replacement as a paired video-to-video translation problem based on paired triplet supervision. During training, each triplet consists of a synthesized driving video, the corresponding original video, and a reference identity image. Instead of reconstructing the synthesized video itself, we reverse the supervision direction by treating the synthesized video as the driving input and the original video as the reconstruction target. Since the paired videos share identical facial performances, body motions, camera trajectories, and scene layouts while differing only in character identity, the model is explicitly supervised to recover the original identity while preserving the complete spatial-temporal dynamics. This formulation provides dense pixel-level supervision over the entire video and enables the model to directly learn identity transformation without relying on intermediate motion representations.

Based on the paired triplet supervision, we perform controllable character replacement using an In-Context Diffusion Transformer built upon Wan2.2. Specifically, the synthesized driving video $V_{\mathrm{syn}}$, original video $V_{\mathrm{gt}}$, reference identity image $I_{\mathrm{ref}}$, and text prompt $P$ are encoded by the frozen Wan2.2 encoders into the driving latent $Z_s$, original latent $Z_o$, reference latent $Z_k$, and text embedding $E_{\mathrm{txt}}$, respectively. We parameter-efficiently fine-tune the pretrained diffusion transformer using Motion LoRA while keeping the backbone frozen. The synthesized driving latent is concatenated with the noisy original latent through channel expansion before patch embedding, while the reference identity latent is independently encoded and concatenated with the video tokens in an in-context manner. The unified token sequence is then processed by the diffusion transformer to transfer the reference identity while preserving the original motion, scene dynamics, and actor interactions.


Following Wan2.2, training is performed under the standard flow-matching formulation. Given the noisy original latent $Z_o^{\tau}$ at diffusion timestep $\tau$, the conditional diffusion transformer predicts the velocity target

\begin{equation}
\hat{v}_{\theta}
=
\mathcal{D}
(
Z_o^{\tau},
\tau;
Z_s,
Z_k,
E_{\mathrm{txt}}
),
\end{equation}

and is optimized using

\begin{equation}
\mathcal{L}_{\mathrm{FM}}
=
\mathbb{E}_{\tau}
\left[
w(\tau)
\|
\hat{v}_{\theta}
-
v_{\tau}
\|_2^2
\right],
\end{equation}

where $v_{\tau}$ denotes the velocity target defined by the flow-matching scheduler and $w(\tau)$ is the timestep-dependent weighting function.

\subsection{3.3 Expression Enhancement Adapter}

Although the proposed controllable video editing framework effectively learns global identity transformation, preserving fine-grained facial expressions remains challenging due to the limited representation capacity of latent video features. This issue becomes more severe in multi-person scenes with different face scales and occlusions.

To address this limitation, we propose an Expression Enhancement Adapter that explicitly models facial expression dynamics while preserving the replacement capability learned in Stage I. Given an aligned facial sequence $F=\{f_1,\ldots,f_T\}$ extracted from the driving video, a frozen Motion Extractor first obtains temporally consistent facial motion features:

\begin{equation}
M=\mathcal{M}(F),
\end{equation}

which are further encoded by a lightweight Face Encoder into expression tokens:

\begin{equation}
E_{\mathrm{exp}}
=
\mathcal{F}(M).
\end{equation}

The expression tokens are injected into the frozen diffusion transformer through Face Adapter modules inserted every fifth DiT block. Specifically, each adapter performs cross-attention between intermediate DiT features and expression tokens, where video features serve as queries and expression tokens provide keys and values. This enables progressive refinement of facial details at multiple semantic levels without altering the global latent representation.

For multi- and selective-character replacement, each actor is associated with an individual facial sequence and corresponding expression tokens before injection. This actor-specific routing prevents expression interference across identities and improves expression preservation in crowded scenes. The actor-specific routing only determines the correspondence between expression tokens and target actors, without introducing additional generation constraints. During this stage, only the Face Encoder and Face Adapter are optimized, while the diffusion backbone, Motion LoRA, and Motion Extractor remain frozen, preserving the replacement capability learned in Stage I.

\begin{figure*}[t]
    \centering
    \includegraphics[width=0.95\textwidth]{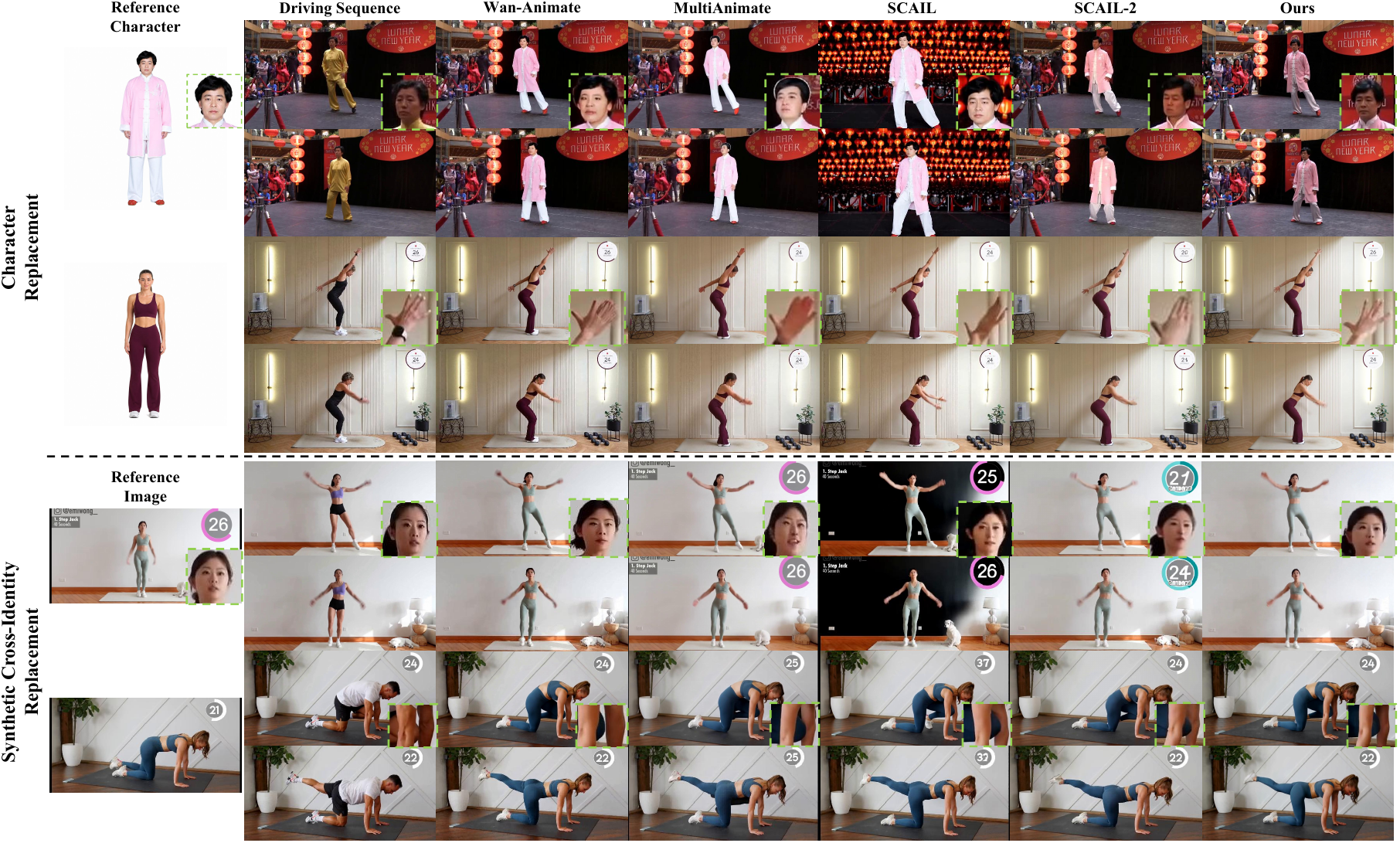}
    \caption{Qualitative comparison on single-character replacement. Zoom in for better visualization.}
    \label{fig:results(1)}
\end{figure*}

\subsection{3.4 Synthetic Data Generation}

Since paired replacement videos with identical performances but different identities are rarely available, we develop a synthetic data generation pipeline to construct large-scale training data. As shown in Fig.~\ref{data}, the pipeline consists of two stages. Stage I synthesizes cross-identity motion pairs for training the base replacement model, while Stage II constructs an expression-stable head dataset for training the Expression Enhancement Adapter. Together, the resulting data enables end-to-end learning of identity transformation while preserving facial expressions, body motions, and scene dynamics.

\textbf{Stage I: Cross-Identity Motion Pair Synthesis.} Starting from web-collected human videos, we synthesize cross-identity pairs using Wan2.2-Animate~\cite{wan2025wan} as a data generator, while preserving motions and scene dynamics. Quality filtering removes failed samples, yielding paired videos with identical performances but different identities. The synthesizer is only used for data construction and is not involved during inference.

\textbf{Stage II: Expression-Stable Head Dataset.} We further construct an expression-stable head dataset from talking-head videos and NeRSemble~\cite{kirschstein2023nersemble} captures. Face tracking, temporal alignment, and quality filtering~\cite{karras2019style} are applied to obtain temporally consistent facial sequences for training the Expression Enhancement Adapter.

\begin{figure*}[t]
    \centering
    \includegraphics[width=0.95\textwidth]{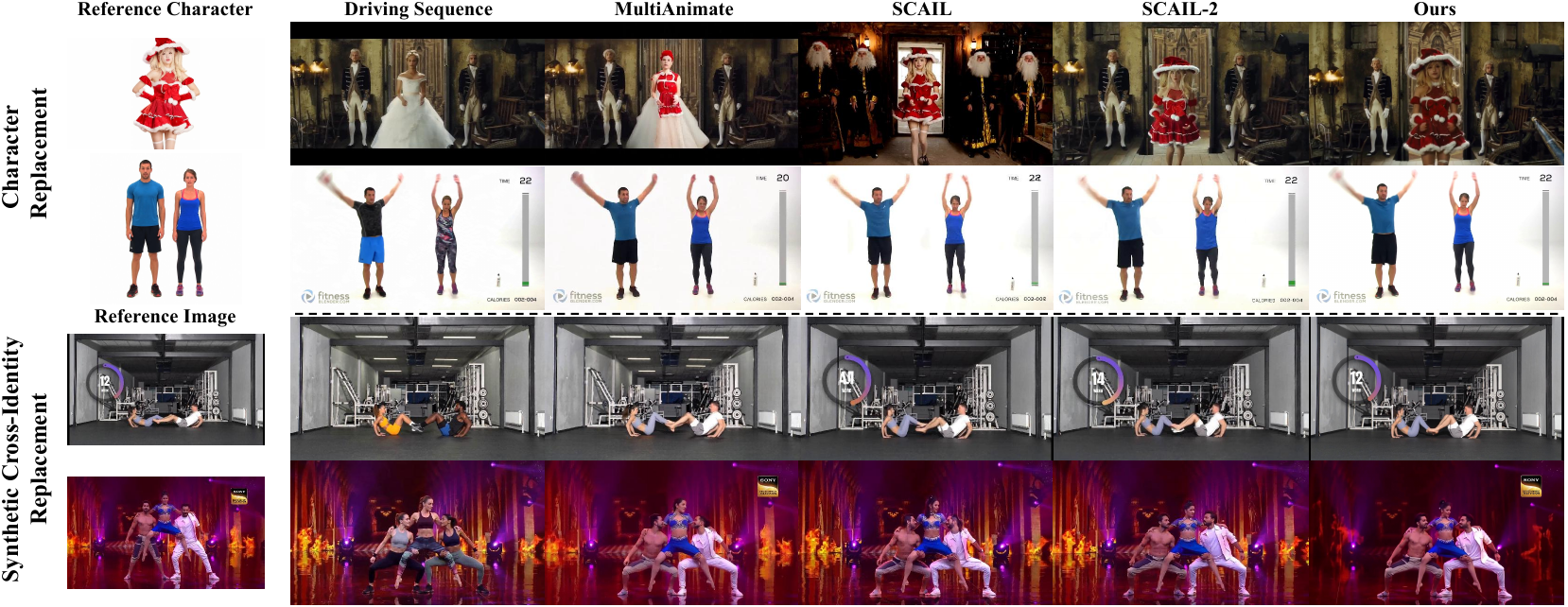} 
    \caption{Qualitative comparison on multi-character replacement. Zoom in for better visualization.}
    \label{fig:results(2)}
\end{figure*}

\begin{table*}[t]
\centering
\setlength{\tabcolsep}{3pt}
\begin{tabular}{llcccccccc}
\toprule
\multirow{2}{*}{Setting} &
\multirow{2}{*}{Method} &
\multicolumn{4}{c}{Motion Preservation} &
\multicolumn{4}{c}{Reconstruction Quality}\\
\cmidrule(lr){3-6}\cmidrule(l){7-10}
&
&
Pose-DTW$\downarrow$ &
PCK@0.2$\uparrow$ &
AJVE$\downarrow$ &
Accel.Err.$\downarrow$ &
PSNR$\uparrow$ &
LPIPS$\downarrow$ &
SSIM$\uparrow$ &
MSE$\downarrow$
\\
\midrule

\multirow{5}{*}{Single-}
& Wan-Animate
&0.0345&0.9824&0.0185&0.0278&18.85&0.2371&0.7947&847.1\\

& MultiAnimate
&0.0387&0.9791&0.0184&0.0281&18.55&0.2380&0.7864&907.7\\

& SCAIL
&0.0298&0.9830&0.0179&0.0283&19.86&0.2323&0.8021&676.1\\

& SCAIL-2
&0.0263&0.9879&0.0177&0.0267&19.89&0.2349&0.8062&666.9\\

& \textbf{ReXMo (Ours)}
&\textbf{0.0258}&\textbf{0.9890}&\textbf{0.0174}&\textbf{0.0265}
&\textbf{19.91}&\textbf{0.2315}&\textbf{0.8075}&\textbf{663.6}
\\

\midrule

\multirow{4}{*}{Multi-}
& MultiAnimate
&0.0428&0.9715&0.0187&0.0284&18.39&0.2394&0.7684&942.3\\

& SCAIL
&0.0341&0.9805&0.0184&0.0292&19.67&0.2336&0.7959&701.4\\

& SCAIL-2
&0.0314&0.9850&0.0183&0.0275&19.68&0.2365&0.7974&700.4\\

& \textbf{ReXMo (Ours)}
&\textbf{0.0312}&\textbf{0.9855}&\textbf{0.0181}&\textbf{0.0271}
&\textbf{19.71}&\textbf{0.2327}&\textbf{0.7987}&\textbf{698.2}
\\

\bottomrule
\end{tabular}

\caption{Quantitative comparison on the synthetic cross-identity replacement benchmark.}
\label{tab:quantitative_results}
\end{table*}

\section{4. Experiments}

\textbf{Dataset}. As shown in Figure~\ref{data}, the Stage-I dataset covers single-person, multi-person, and talking-face scenarios. Each sample contains a synthesized replacement video and its original counterpart with matched motion and scene dynamics. The synthesized data is used only for training, while evaluation is performed on held-out videos. Stage-II uses approximately 30K aligned facial clips from YouTube and NeRSemble~\cite{kirschstein2023nersemble}, processed into temporally consistent $512\times512$ expression sequences.

\textbf{Training Setting}. We build our framework on the Wan2.2-TI2V-5B diffusion backbone and train it in two stages. Stage I performs rank-80 Motion LoRA fine-tuning for reference-guided character replacement on the paired multi-resolution video dataset, while freezing the pretrained VAE, text encoder, and DiT backbone. The expanded patch embedding layer for condition-video injection is jointly optimized. Stage II introduces the Expression Enhancement Adapter, where aligned facial-expression sequences are encoded into expression tokens through a frozen Motion Extractor and a lightweight Face Encoder. These tokens are injected through Face Adapter modules every five DiT blocks, with only the Face Encoder and Face Adapter optimized while other components remain frozen. Both stages are trained on 97-frame clips at 24 FPS for 10k iterations using AdamW with bf16 mixed precision, gradient checkpointing, and a learning rate of $5\times10^{-5}$. Additional implementation details are provided in the supplementary material.


\textbf{Evaluation Metrics and Baselines}. We compare ReXMo with four open-source baselines, including Wan-Animate~\cite{wan2025wan}, MultiAnimate~\cite{hu2026multianimate}, SCAIL~\cite{yan2026scail}, and SCAIL-2~\cite{yan2026scail2}. We establish two evaluation protocols: 1) \textbf{Character Replacement}. Since paired ground-truth videos are unavailable, we follow previous work and adopt Video-Bench~\cite{han2025video} for perceptual evaluation, reporting Imaging Quality, Motion Smoothness, Temporal Consistency, and Appearance Consistency. We evaluate both single- and multi-character replacement scenarios. Wan-Animate is included only in the single-character setting due to its focus on single-character animation. 2) \textbf{Synthetic Cross-Identity Replacement}. We construct a synthetic paired benchmark with paired correspondence for quantitative evaluation. Given an original video $A$, a cross-identity video $B$ is generated using a reference identity, and the original identity is reconstructed as $\bar{A}$ using a reference frame from $A$. We compare $\bar{A}$ with $A$ using motion metrics (Pose-DTW, PCK@0.2, AJVE, and Accel.Err.) and identity recovery metrics (PSNR, SSIM, LPIPS, and MSE). Facial expression preservation is further evaluated using landmark-based metrics, including F-LMD, M-LMD, Expr-LMD, and LVD. Results are reported separately for single- and multi-character scenarios.

\subsection{4.1 Results and Comparisons}

\textbf{Quantitative Comparison}. Table~\ref{tab:quantitative_results} presents quantitative comparisons on the synthetic cross-identity replacement benchmark under single- and multi-character settings. Given an original video, each method first performs cross-identity replacement using a reference identity and then reconstructs the original identity from the generated video. The reconstructed videos are evaluated against the original videos using motion preservation and reconstruction quality metrics. Our method achieves the best overall performance across both evaluation settings. Compared with previous methods, it consistently improves motion preservation, obtaining the lowest Pose-DTW and Accel. Err., while achieving the highest PCK@0.2 in both single- and multi-character scenarios. Although SCAIL-2~\cite{yan2026scail2} provides competitive motion consistency, our method further improves identity recovery quality with higher PSNR and SSIM and lower LPIPS and MSE, demonstrating more faithful identity recovery and appearance preservation. 


We further evaluate generated videos using Video-Bench~\cite{han2025video} across four dimensions: Image Quality, Motion Smoothness, Temporal Consistency, and Appearance Consistency. As shown in Table~\ref{tab:video_bench}, ReXMo achieves the best performance in three dimensions, demonstrating superior visual quality and temporal and appearance consistency.

\begin{table}[t]
\centering
\setlength{\tabcolsep}{1.5pt}
\begin{tabular}{lcccc}
\toprule
Method & 
\makecell[c]{Image\\Quality$\uparrow$} & 
\makecell[c]{Motion\\Smooth.$\uparrow$} & 
\makecell[c]{Temporal\\Consist.$\uparrow$} & 
\makecell[c]{Appearance\\Consist.$\uparrow$} \\
\midrule
Wan-Animate  & 3.78 & 3.88 & 3.93 & 4.18 \\
MultiAnimate & 3.71 & 3.74 & 3.84 & 3.90 \\
SCAIL        & 4.18 & \textbf{3.92} & 3.98 & 4.20 \\
SCAIL-2      & 4.27 & 3.87 & 4.08 & 4.32 \\
\textbf{ReXMo (Ours)} 
             & \textbf{4.32} & 3.90 & \textbf{4.12} & \textbf{4.39} \\
\bottomrule
\end{tabular}
\caption{Perceptual comparison on the character replacement.}
\label{tab:video_bench}
\end{table}

\begin{figure}[t]
    \centering
    \includegraphics[width=0.46\textwidth]{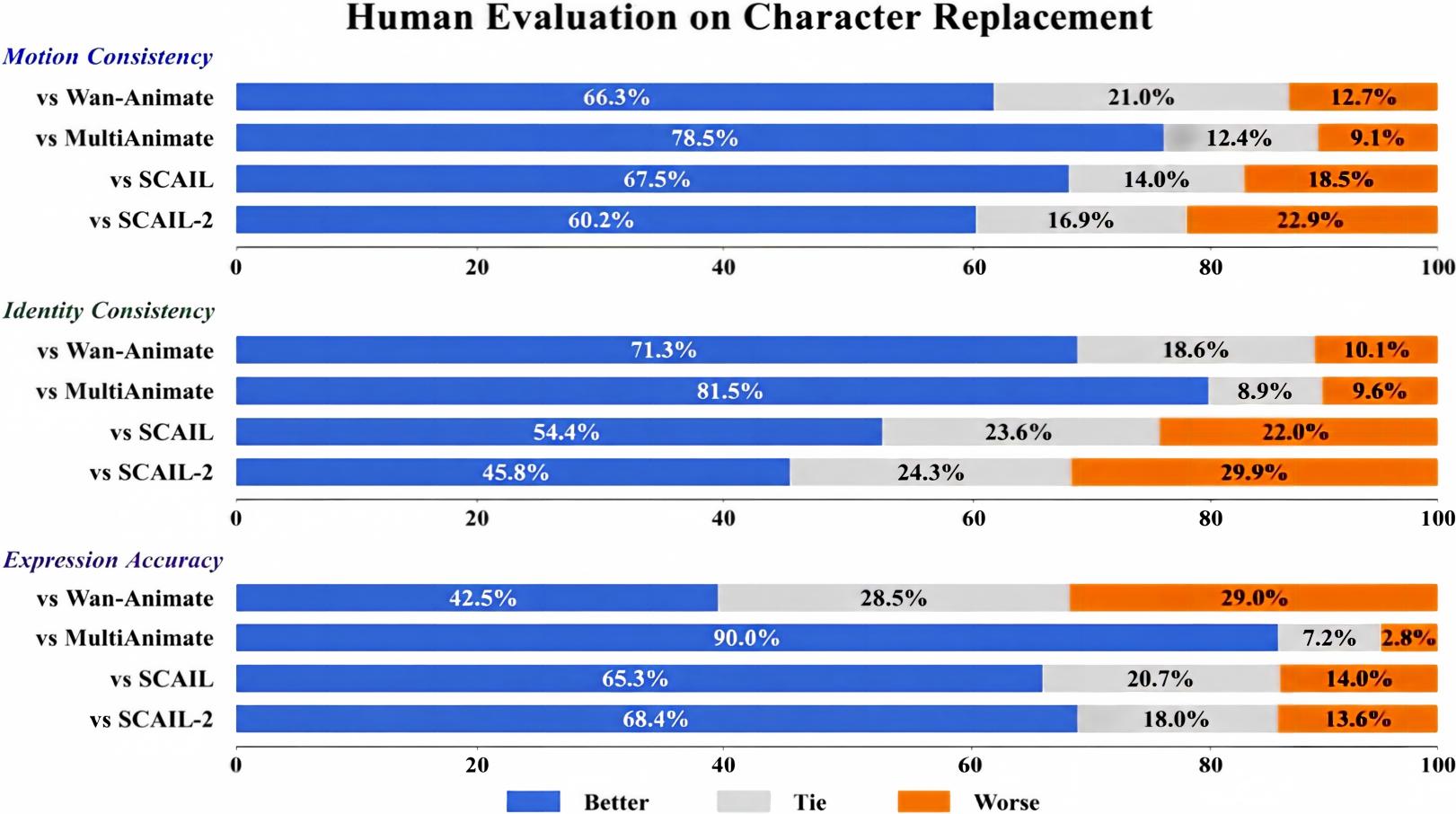} 
    \caption{User Study (50 participants) on Motion Consistency, Identity Consistency, and Expression Accuracy.}
    \label{fig:user_study}
\end{figure}

\begin{table}[t]
\centering
\setlength{\tabcolsep}{1.8pt}
\begin{tabular}{lcccc}
\toprule
Method & F-LMD$\downarrow$ & M-LMD$\downarrow$ & Expr-LMD$\downarrow$ & LVD$\downarrow$ \\
\midrule
Wan-Animate & 0.3345 & 0.3248 & 0.3189 & 0.0374 \\
MultiAnimate & 0.5013 & 0.4977 & 0.5086 & 0.0458 \\
SCAIL & 0.3309 & 0.3381 & 0.3485 & 0.0388 \\
SCAIL-2 & 0.3350 & 0.3411 & 0.3414 & 0.0394 \\
\midrule
w/o EEA & 0.3341 & 0.3251 & 0.3415 & 0.0389 \\
\textbf{w/ EEA} & \textbf{0.2858} & \textbf{0.2929} & \textbf{0.2905} & \textbf{0.0288} \\
\bottomrule
\end{tabular}
\caption{Quantitative ablation study of the Expression Enhancement Adapter.}
\label{tab:expr_ablation}
\end{table}

\begin{figure}[t]
    \centering
    \includegraphics[width=0.45\textwidth]{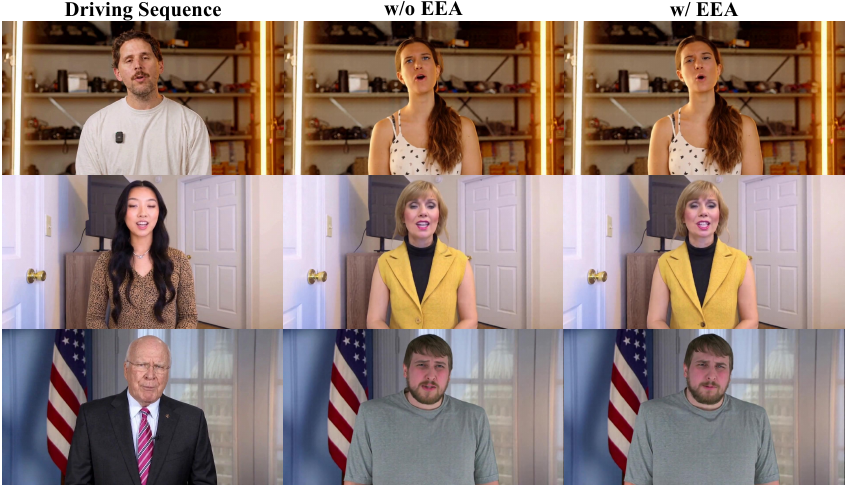} 
    \caption{Visualization of the ablation study on the Expression Enhancement Adapter (EEA). w/ and w/o EEA denote models with and without the proposed adapter.}
    \label{fig:ablation}
\end{figure}

\textbf{Qualitative Comparison}. We present qualitative comparisons with existing methods under character replacement and synthetic cross-identity replacement settings. For single-character replacement, as shown in Fig.~\ref{fig:results(1)}, ReXMo preserves both identity characteristics and original performance details more faithfully. Compared with pose-driven approaches, our method better maintains fine-grained facial expressions, subtle body motions, and appearance consistency with the reference identity. In challenging cases with large motion variations, previous methods tend to suffer from identity drift, blurred facial details, or expression inconsistency, while ReXMo maintains more stable identity and expression preservation. For multi-character and selective replacement, Fig.~\ref{fig:results(2)} demonstrates that ReXMo effectively handles multiple identities while preserving individual motion patterns and expression consistency. Existing methods may introduce identity confusion or expression interference when multiple actors interact, leading to degraded facial details and inconsistent appearances. By extracting and enhancing facial features for different actors, the Expression Enhancement Adapter enables more accurate expression preservation in complex multi-person scenes.

Furthermore, the synthetic cross-identity results show that ReXMo preserves both identity and motion consistency, validating the effectiveness of our paired triplet supervision.

\subsection{4.2 User Study}

We conduct a user study with 50 participants to evaluate perceptual preferences among different methods. Participants compare ReXMo with each baseline under three criteria: Motion Consistency, Identity Consistency, and Expression Accuracy. Each comparison pair is randomly shuffled to avoid positional bias, and participants select whether ReXMo performs better, ties, or worse based on overall visual quality. The pairwise comparison results are shown in Fig.~\ref{fig:user_study}.


\subsection{4.3 Ablation Study}
We conduct an ablation study to evaluate the effectiveness of the Expression Enhancement Adapter (EEA). As shown in Table~\ref{tab:expr_ablation}, removing EEA results in consistent performance degradation on all expression-related metrics. Specifically, EEA reduces F-LMD from 0.3341 to 0.2858, Expr-LMD from 0.3415 to 0.2905, and LVD from 0.0389 to 0.0288, demonstrating more accurate expression alignment and improved temporal expression consistency. These improvements indicate that the latent video representation learned during Stage I is insufficient to fully preserve subtle facial dynamics, while the proposed adapter provides complementary expression-aware guidance for fine-grained facial motion reconstruction.

Qualitative comparisons are shown in Fig.~\ref{fig:ablation}, where the driving video, the results without EEA, and the full ReXMo model are compared. Without EEA, the generated videos tend to lose subtle facial movements, such as mouth deformation and expression intensity variations, resulting in weaker correspondence with the driving performance. In contrast, EEA preserves fine-grained facial dynamics and produces expressions that are more consistent with the input driving video.

These results verify that decoupling identity replacement and expression enhancement enables ReXMo to achieve faithful facial expression preservation while maintaining motion and scene consistency during character replacement.

\section{5. Conclusion}

In this work, we present ReXMo, an end-to-end framework for high-fidelity video character replacement with expression and motion preservation. By introducing paired triplet supervision and an In-Context Diffusion Transformer, ReXMo learns identity transformation while preserving motion, scene dynamics, and interactions. The proposed Expression Enhancement Adapter further improves fine-grained expression preservation, enabling high-quality character replacement across diverse scenarios. Extensive experiments demonstrate superior performance in motion consistency, identity fidelity, visual quality, and expression preservation.


\bibliography{aaai2027}


\end{document}